\documentclass[conference]{IEEEtran}
\usepackage{times}

\usepackage[numbers]{natbib}
\usepackage{multicol}
\usepackage[bookmarks=true]{hyperref}

\usepackage{graphicx} 
\usepackage{xcolor} 
\usepackage{array} 
\usepackage{multirow} 
\usepackage{tikz}   
\usepackage{amsmath}   
\usetikzlibrary{arrows.meta, positioning, calc}   

\usepackage[T1]{fontenc}
\usepackage{textcomp}
\makeatletter
\newcommand{\conf}[1]{%
  \def\ps@IEEEtitlepagestyle{%
    \def\@oddhead{%
      \parbox[b]{\dimexpr \textwidth}{\centering\normalsize #1}
    }%
    \let\@evenhead\@empty
    \let\@oddfoot\@empty
    \let\@evenfoot\@empty
  }%
}
\makeatother
\conf{RSS’24 Workshop on Semantics for Robotics: From Environment Understanding and Reasoning to Safe Interaction}

\begin{document}

\title{Exploiting Radiance Fields for Grasp Generation on Novel Synthetic Views}


\author{\authorblockN{Abhishek Kashyap}
\authorblockA{Centre for Applied Autonomous\\ Sensor Systems (AASS)\\
Örebro University\\
Örebro, Sweden 701 82\\
Email: abhishek.kashyap@oru.se}
\and
\authorblockN{Henrik Andreasson}
\authorblockA{Centre for Applied Autonomous\\ Sensor Systems (AASS)\\
Örebro University\\
Örebro, Sweden 701 82\\
Email: henrik.andreasson@oru.se}
\and
\authorblockN{Todor Stoyanov}
\authorblockA{Centre for Applied Autonomous\\ Sensor Systems (AASS)\\
Örebro University\\
Örebro, Sweden 701 82\\
Email: todor.stoyanov@oru.se}}


%


\maketitle

\begin{abstract}
Vision based robot manipulation uses cameras to capture one or more images of a scene containing the objects to be manipulated. Taking multiple images can help if any object is occluded from one viewpoint but more visible from another viewpoint. However, the camera has to be moved to a sequence of suitable positions for capturing multiple images, which requires time and may not always be possible, due to reachability constraints. So while additional images can produce more accurate grasp poses due to the extra information available, the time-cost goes up with the number of additional views sampled. Scene representations like Gaussian Splatting are capable of rendering accurate photorealistic virtual images from user-specified novel viewpoints. In this work, we show initial results which indicate that novel view synthesis can provide additional context in generating grasp poses. Our experiments on the Graspnet-1billion dataset show that novel views contributed force-closure grasps in addition to the force-closure grasps obtained from sparsely sampled real views while also improving grasp coverage. In the future we hope this work can be extended to improve grasp extraction from radiance fields constructed with a single input image, using for example diffusion models or generalizable radiance fields.
\end{abstract}

\IEEEpeerreviewmaketitle

\section{Introduction}
RGB-D cameras are widely used in the robotics community to obtain visual information of a robot’s surroundings. The perception data acquired from the camera has to be processed to create a representation of the robot’s environment for the intended
downstream task, examples being navigation and manipulation. In the field of robot manipulation, a 3D scene representation has applications like object detection, classification, segmentation, pose estimation, and grasp planning. A scene representation which allows deriving geometric and semantic context of a robot's environment can be valuable for environment understanding and can be combined with other sensory inputs like tactile sensors or force sensors to provide multi-modal information.

Creating a scene representation usually requires capturing multiple images of the scene from distinct viewpoints. While more images can be useful in providing more perspectives, this comes with the time-cost of moving the camera to all the desired viewpoints the scene has to be observed from. For a robot manipulator with an eye-in-hand camera, the number of image capture viewpoints and their spatial distribution may require large robot motions. Constructing a scene representation with as few viewpoints as possible while at the same time having the ability to observe the scene from more viewpoints than were used to construct the scene representation would be advantageous.

Radiance fields like Neural Radiance Fields (NeRFs)~\citep{mildenhall2021nerf} and Gaussian Splatting~\citep{kerbl20233d} have shown a remarkable ability to synthesize novel views of a scene. These novel views can serve as extra viewpoints to observe the scene and help acquire more context of the scene than would be possible when only real views are available. In this work, we explore the usefulness of novel views for grasp generation. Our hypothesis is that given a radiance field, acquiring renders from novel viewpoints can provide additional useful context for generating grasp poses. The main contributions of this paper are:
    \begin{itemize}
    \item Demonstrating how novel view synthesis can produce force-closure grasp poses in addition to poses obtained from real viewpoints.
    \item Showing how novel view synthesis can increase grasp coverage in the scene which is the number of objects in the scene for which valid grasp poses could be computed.
    \end{itemize}

This paper is organized as: section \ref{section_background} provides background on the topic of scene representations, section \ref{section_implementation} describes implementation details related to proving the hypothesis, section \ref{section_experiments_results} summarizes the findings of this study, and section \ref{section_conclusion} includes concluding remarks and ways to extend this work.

\section{Background}
\label{section_background}
Scene representations that have been used for robot manipulation include pointclouds~\citep{ten2017grasp, liang2019pointnetgpd, mousavian20196, fang2020graspnet, sundermeyer2021contact}, meshes~\citep{lundell2020beyond, agnew2021amodal}, voxels~\citep{wada2017probabilistic, wada2020morefusion, james2022coarse, shridhar2023perceiver}, signed distance fields~\citep{stoyanov2016grasp, song2020grasping, breyer2021volumetric}, and neural radiance fields~\citep{ichnowski2021dex, jiang2021synergies, kerr2022evo, dai2023graspnerf, soti2023gradient, liu2024rgbgrasp, soti20246}. Representations like point clouds, meshes, and voxels are explicit representations because the representation geometrically fits surfaces and occupied volumes in the scene. Implicit representations like signed distance fields and neural radiance fields on the other hand parameterize the scene using a continuous function that can be queried to obtain information at a specific location in the scene.

A neural radiance field (NeRF) encodes the scene in the weights of a fully-connected neural network which takes as input a position and a viewing direction, and outputs the color and density at that position~\citep{mildenhall2021nerf}. An additional feature of NeRF is its ability to produce highly photorealistic renders from novel viewpoints. High quality novel-view-synthesis is also achievable through a more recent work that optimizes Gaussians to encode scene information and then renders views by projecting the Gaussians on to the viewpoint's 2D plane~\citep{kerbl20233d}. The projection (``splatting'') is parallelized using a tile-based rasterizer making it faster than volume ray casting methods. Gaussian Splatting has begun to see widespread usage in scene representations~\citep{chen2024survey}, which in turn are used for downstream tasks like Simultaneous Localization and Mapping (SLAM)~\citep{tosi2024nerfs} and robot manipulation~\citep{lu2024manigaussian, zheng2024gaussiangrasper}. While there has been previous work on using radiance fields for manipulation~\citep{ichnowski2021dex, jiang2021synergies, kerr2022evo, dai2023graspnerf, soti2023gradient, liu2024rgbgrasp, soti20246}, none to the best of our knowledge have sought to leverage novel viewpoints in the context of grasp generation.

\section{Implementation}
\label{section_implementation}
To test our hypothesis, we use the Graspnet-1billion dataset~\citep{fang2020graspnet} which is a collection of 190 tabletop scenes, each scene having a random assortment of objects accompanied with RGB-D images from 256 viewpoints on a quarter sphere.

For a scene, a radiance field is created out of $M=3$ images using the technique of Gaussian Splatting~\citep{kerbl20233d}, which then is used to render the tabletop scene from $N=16$ novel viewpoints. Graspnet-1billion's pre-trained grasp detection network is run on the real views used to create the radiance field and also on the synthesized novel views.

The number of force-closure grasps from real and novel views are compared against the number of force-closure grasps obtained from real views. Additionally, grasp coverage is also compared where grasp coverage is expressed as a percentage and represents the number of objects in the scene for which there is at least one force-closure grasp configuration out of all the objects in the scene.

\subsection{Scene representation using Gaussian Splatting}
\label{subsection_impl_scene_rep_3dgs}
We use the implementation from SplaTAM~\citep{keetha2023splatam} to optimize Gaussians which can render color and depth images from novel viewpoints. Meant for dense RGB-D Simultaneous Localization and Mapping (SLAM), SplaTAM performs three steps for every new RGB-D frame: \textit{Camera Tracking} to estimate camera pose of the new frame, \textit{Gaussian Densification} to initialize new Gaussians in the scene based on a computed densification mask, and \textit{Map Update} to optimize all Gaussians in the scene by minimizing RGB and depth errors across all keyframe images. We found SplaTAM useful for reconstructing tabletop scenes like those found in the Graspnet-1billion dataset~\citep{fang2020graspnet}. Since camera poses are already available in Graspnet-1billion, camera tracking is not necessary.

\subsection{Selecting real views}
\label{subsection_impl_real_views}
\begin{figure}[htbp]
    \centering
    \includegraphics[width=0.80\columnwidth]{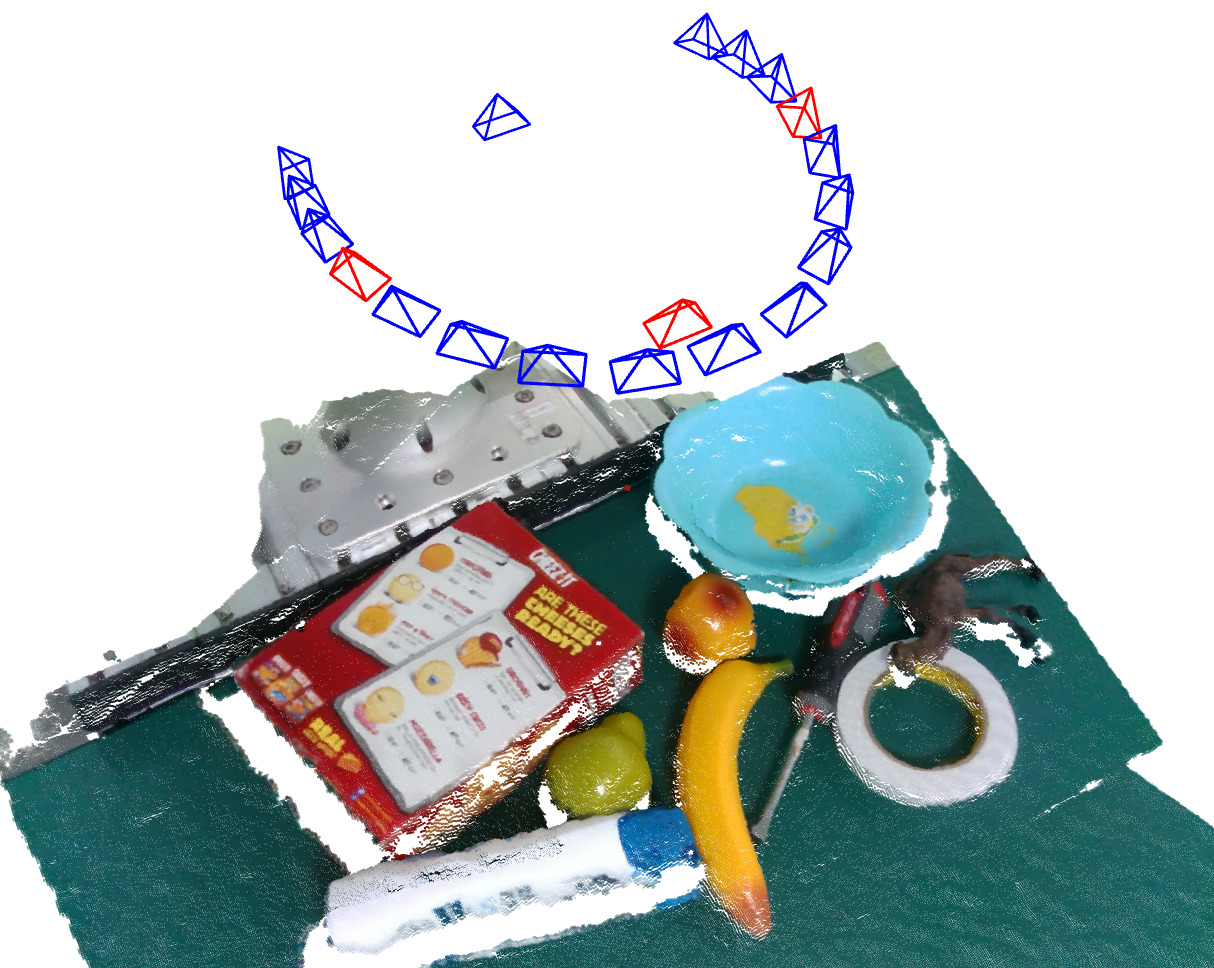}
    \caption{Example scene reconstruction showing camera poses as frustums, all pointing down: real viewpoints in \textcolor{red}{red}, novel viewpoints in \textcolor{blue}{blue}}
    \label{fig:cam_viewpoints}
\end{figure}

\begin{figure}[htbp]
    \centering
    \begin{tabular}{ccc}
        \includegraphics[width=0.28\columnwidth]{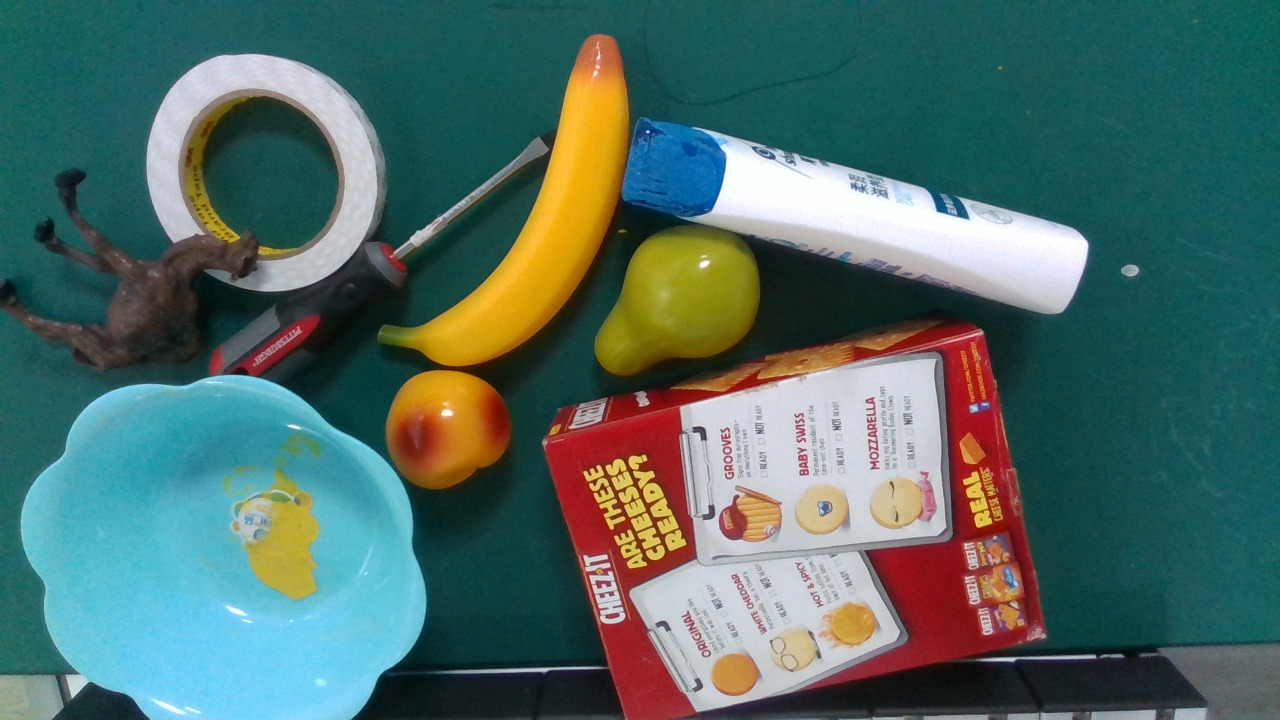} & 
        \includegraphics[width=0.28\columnwidth]{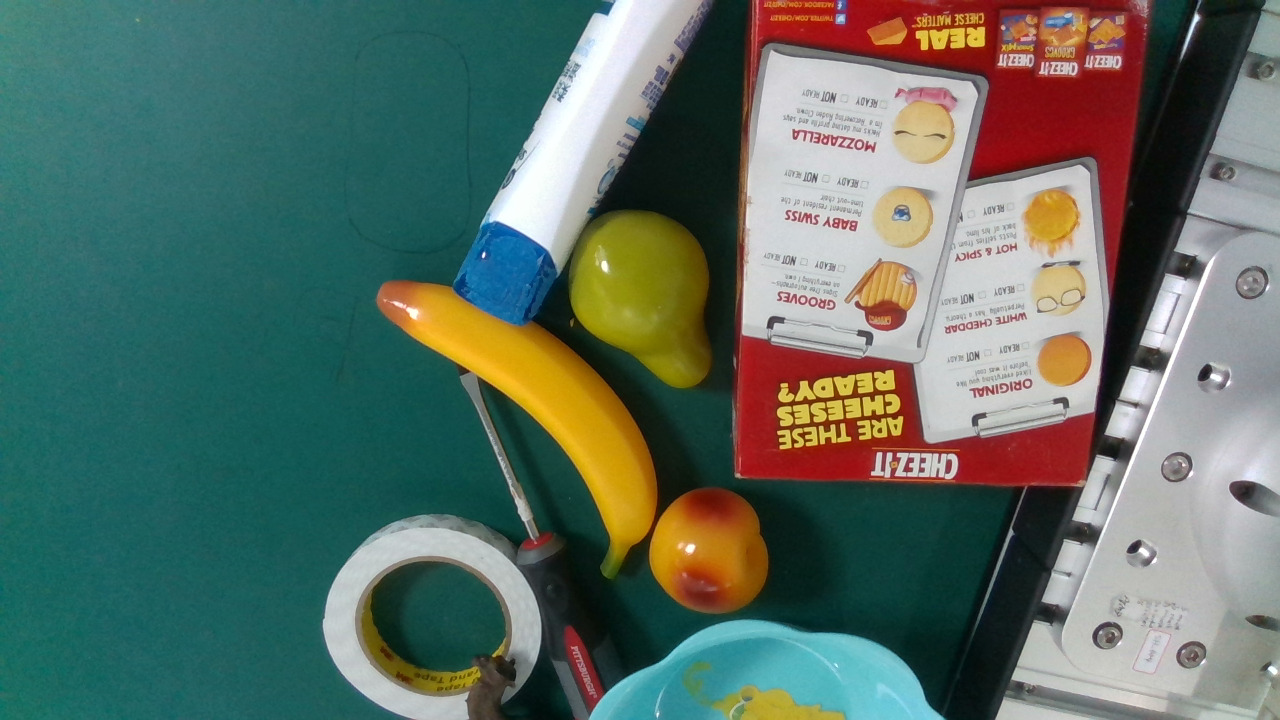} & 
        \includegraphics[width=0.28\columnwidth]{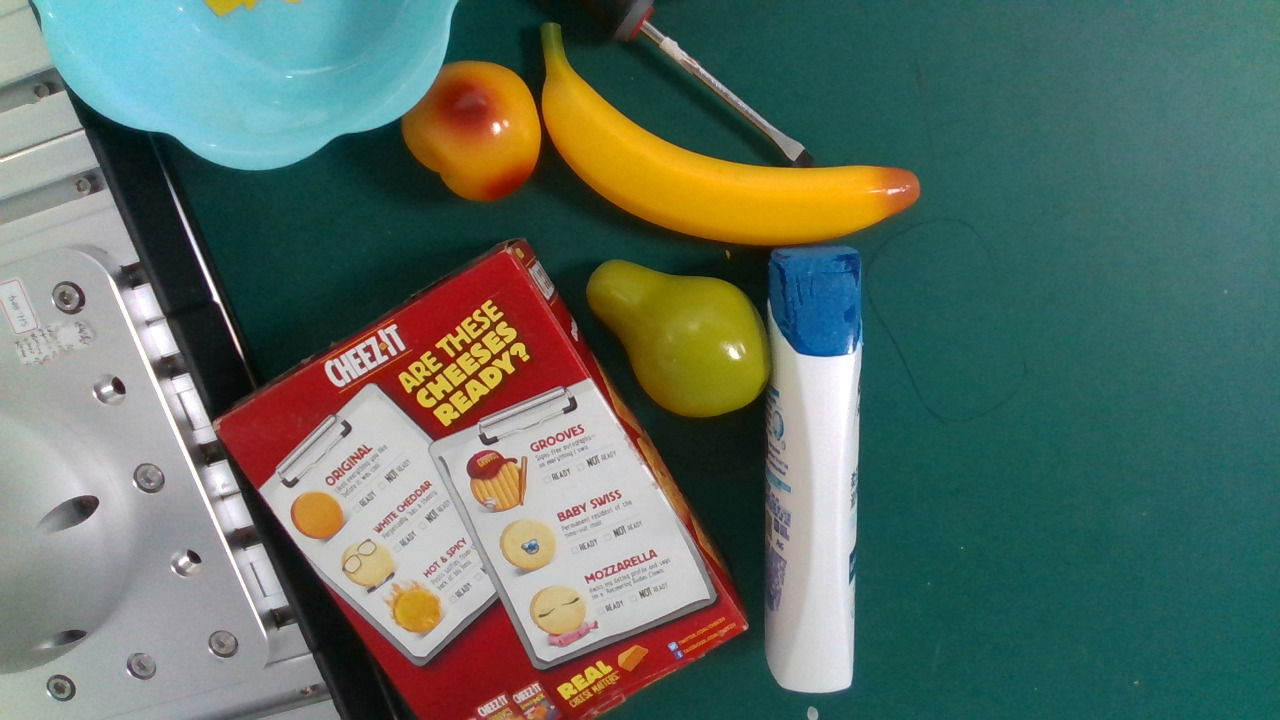} \\
    \end{tabular}
    \caption{Example real views of a scene for constructing a radiance field}
    \label{fig:three_example_real_views}
\end{figure}

Each scene in the Graspnet-1billion dataset has 256 RGB-D images sampled on a quarter sphere. $M=3$ viewpoints of a scene, shown as red frustums in Fig.~\ref{fig:cam_viewpoints}, were selected to create the radiance field representing the scene. Fig.~\ref{fig:three_example_real_views} shows how an example scene looks like from the $M$ viewpoints. We select these views in order to simulate a real-world use case scenario, where the robot has an eye-in-hand camera and the objective is to minimize the motion necessary prior to grasp selection.

\subsection{Selecting novel views}
\label{subsection_novel_views}
The Gaussian Splatting scene reconstruction is used to render color and depth images from $N=16$ novel viewpoints, shown as blue frustums in Fig.~\ref{fig:cam_viewpoints}.
These viewpoints are close to and have similar orientation to the real views because we expected the pre-trained network to produce better quality grasp poses with views familiar from training.

\subsection{Grasp inference}
\label{subsection_impl_grasp_inference}
The Gaussian Splatting based scene reconstruction was projected onto the $M$ real and $N$ novel viewpoints described in sections \ref{subsection_impl_real_views} and \ref{subsection_novel_views} respectively. The scene reconstructions were of high quality, elaborated further in Section\ref{subsection_recon_results}, so projections on the real viewpoints were comparable to the original color and depth images at the real viewpoints.

Projecting the optimized Gaussians generated color and depth images which were used to create point clouds. Grasps were inferred for the $M$ real view point clouds and $N$ novel view point clouds using Graspnet-1billion's pre-trained grasp detection network.

\subsection{Grasp quality metric: force-closure}
\label{subsection_fc_metric}
We use the implementation from Dex-net 2.0~\citep{mahler2017dex} to compute whether a grasp pose achieves force-closure~\citep{nguyen1988constructing}, a binary label which can be either \textit{true} or \textit{false}. Five different coefficients of static friction $\mu$ are used for every inferred grasp to check whether the grasp has achieved force-closure, where $\mu \in \{0.2, 0.4, 0.6, 0.8, 1.0\}$. If force-closure is achieved with any $\mu$, the grasp is reported as having achieved force-closure.

\section{Experiments and Results}
\label{section_experiments_results}

\subsection{Scene reconstruction using Gaussian Splatting}
\label{subsection_recon_results}
SplaTAM uses four metrics for assessing reconstruction quality, three for color and one for depth. Color rendering is assessed with Peak Signal to Noise Ratio (PSNR)↑, Multi-Scale Structural Similarity Index Measure (MS-SSIM)↑~\citep{wang2003multiscale}, and Learned Perceptual Image Patch Similarity (LPIPS)↓~\citep{zhang2018unreasonable}, and depth rendering is assessed with Depth L1 loss↓, with the arrows indicating whether higher or lower is better.

The scene reconstructions exhibited high PSNR and MS-SSIM, low LPIPS, and acceptably low Depth L1 loss, with the average across 190 scenes being 30.608, 0.984, 0.053, and 0.105 respectively. Table.~\ref{tab:compare_rendered_vs_gt}
 shows a comparison of rendered RGB images against ground truth images for 3 out of the $N=16$ novel viewpoints.
\begin{table}[htbp]
    \caption{Comparison of rendered RGB images against ground truth images at novel viewpoints}
    \label{tab:compare_rendered_vs_gt}
    \centering
    \resizebox{0.9\width}{!}{ 
    \begin{tabular}{|>{\centering\arraybackslash}p{0.45\columnwidth}|>{\centering\arraybackslash}p{0.45\columnwidth}|}
        \hline
        Rendered & Ground truth \\
        \hline
        \vspace{0.5mm} \includegraphics[width=0.45\columnwidth]{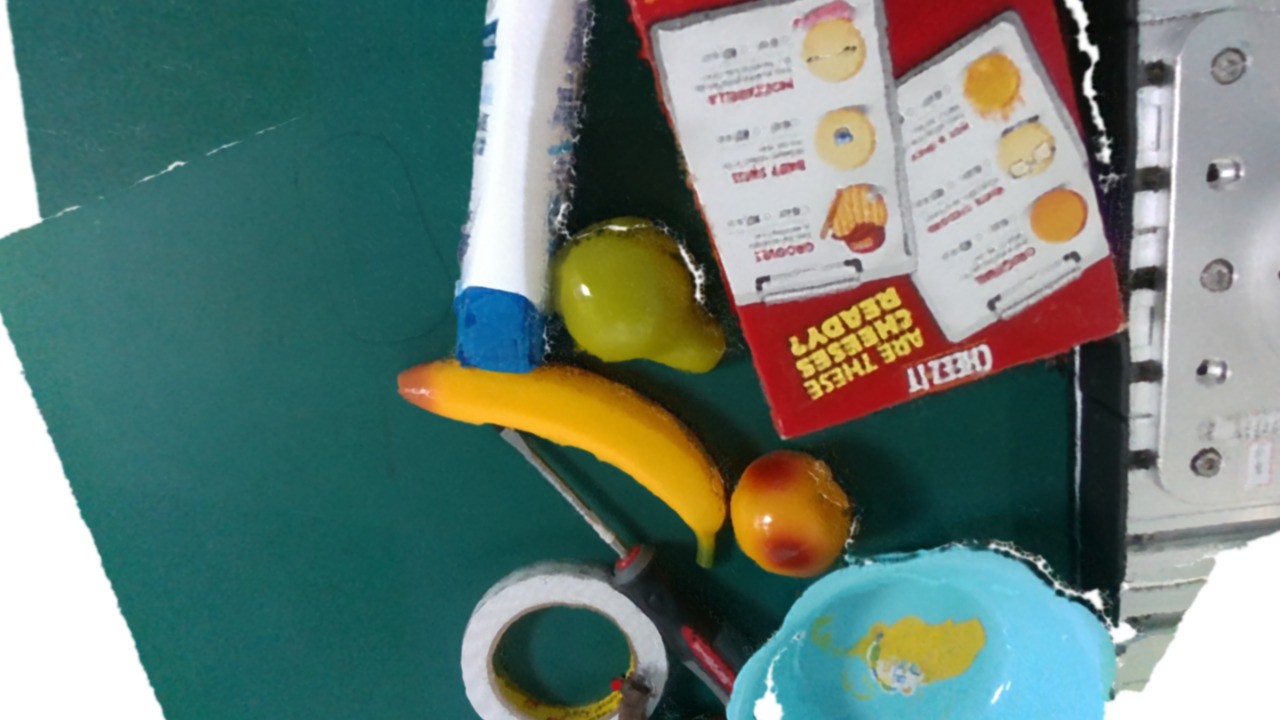} \vspace{0.005mm} & \vspace{0.5mm} \includegraphics[width=0.45\columnwidth]{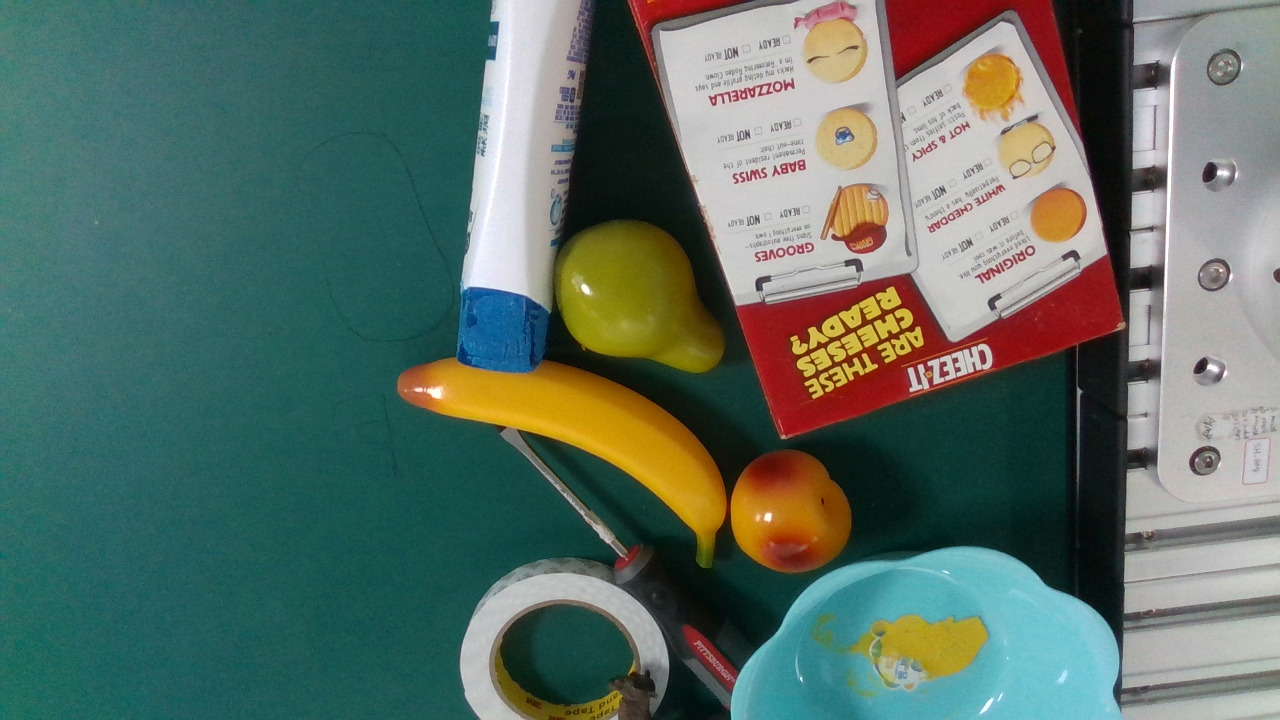} \vspace{0.005mm} \\
        \hline
        \vspace{0.5mm} \includegraphics[width=0.45\columnwidth]{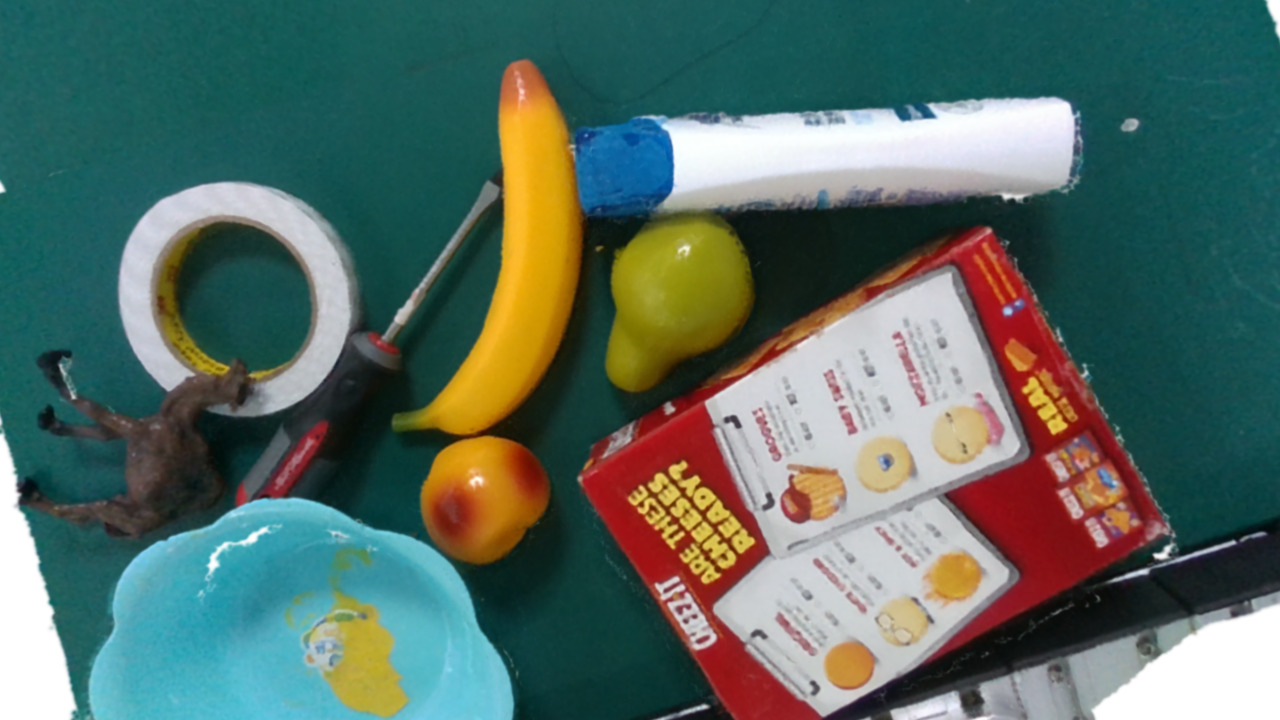} \vspace{0.005mm} & \vspace{0.5mm} \includegraphics[width=0.45\columnwidth]{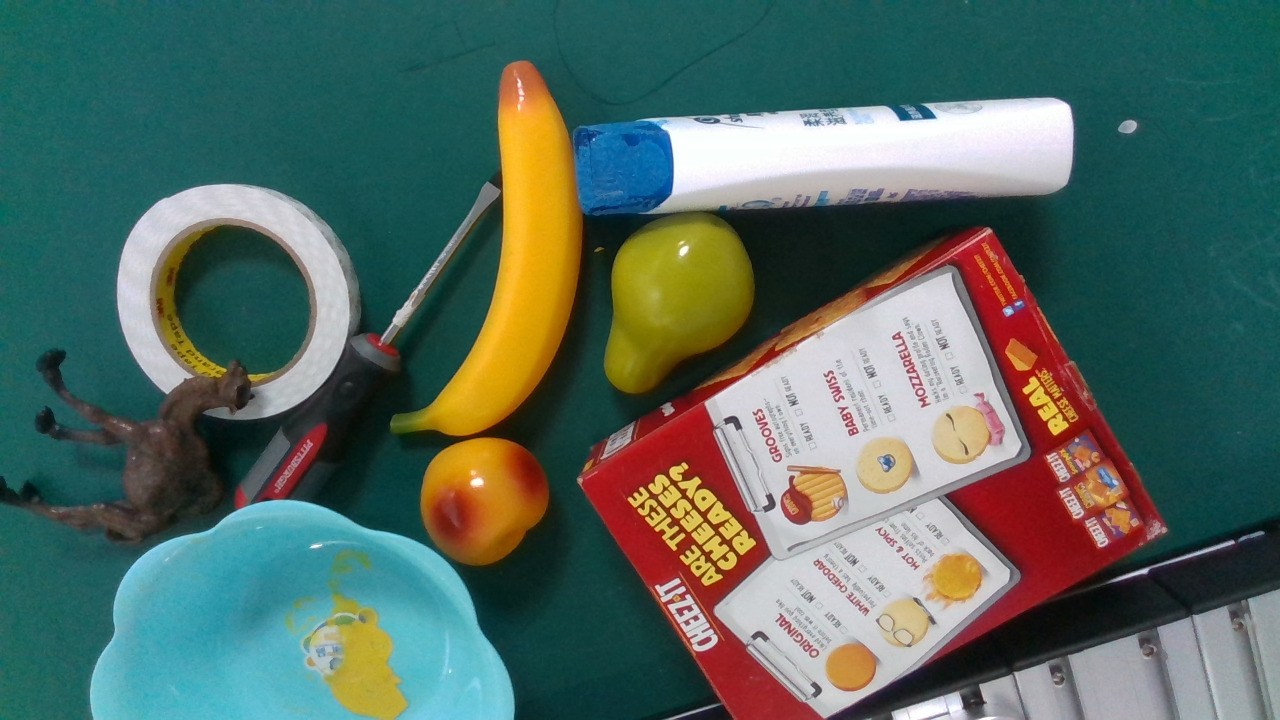} \vspace{0.005mm} \\
        \hline
        \vspace{0.5mm} \includegraphics[width=0.45\columnwidth]{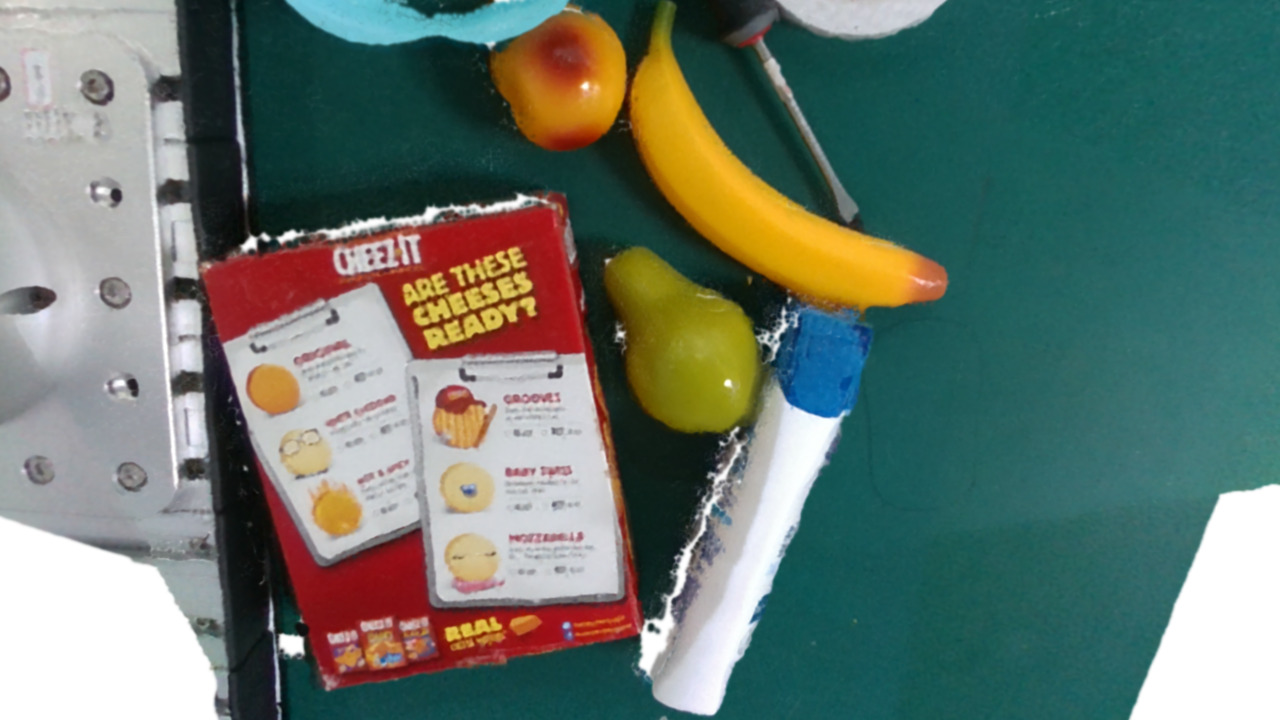} \vspace{0.005mm} & \vspace{0.5mm} \includegraphics[width=0.45\columnwidth]{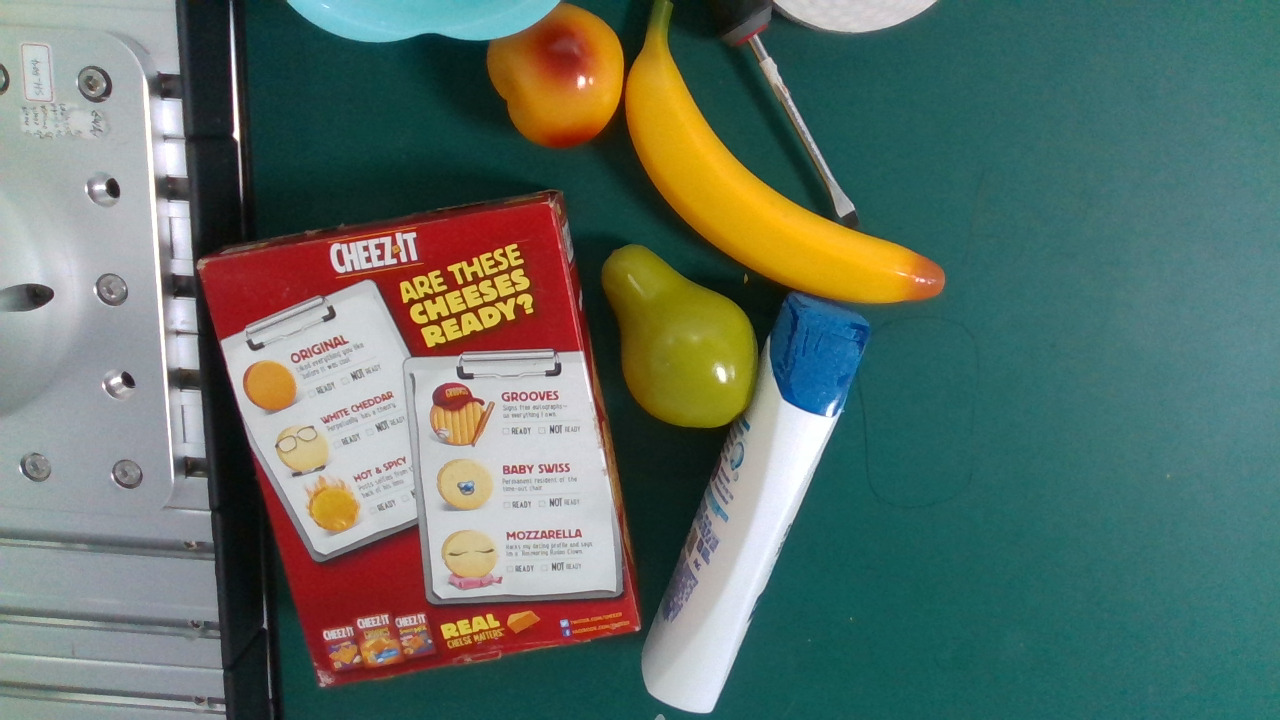} \vspace{0.005mm} \\
        \hline
    \end{tabular}
    }   
\end{table}

\subsection{Grasp generation, aggregation, and post-processing}
\label{subsection_grasp_gen_aggregate_postprocess}

As described in Section~\ref{subsection_impl_grasp_inference}, grasps are inferred from $M$ real views and $N$ novel views, hereafter referred to as $G_{\text{real}}$ and $G_{\text{nvs}}$ respectively (\textit{nvs} refers to novel-view-synthesis). These inferred grasps are then combined into $G_{\text{real+nvs}}$ while still preserving the original grasps from the real and novel views.

We adopted three parallel and independent post-processing branches before performing evaluation: \textbf{i)} apply pose-Non-Maximum-Suppression (pose-NMS) on $G_{\text{real}}$, $G_{\text{nvs}}$, and $G_{\text{real+nvs}}$, \textbf{ii)} apply clustering and top-grasp filtering on $G_{\text{real}}$, $G_{\text{nvs}}$, and $G_{\text{real+nvs}}$, and \textbf{iii)} keep $G_{\text{real}}$, $G_{\text{nvs}}$, and $G_{\text{real+nvs}}$ as is.

\subsubsection{Application of pose-NMS}
\label{subsubsection_pose_nms}
Similar to~\citet{fang2020graspnet}, pose-NMS is applied on the grasps which merges every pair of grasps that have their translation and rotation distance under specified thresholds. This operation reduces redundant grasp poses as may happen when aggregating grasps from multiple perspectives. The default thresholds from Graspnet-1billion are used: translation distance of $0.03m$ and rotation distance of $15^{\circ}$. Applying pose-NMS on $G_{\text{real}}$, $G_{\text{nvs}}$, and $G_{\text{real+nvs}}$ produces $G_{\text{nms(real)}}$, $G_{\text{nms(nvs)}}$, and $G_{\text{nms(real+nvs)}}$. Note that $G_{\text{nms(real+nvs)}}$ is no longer the sum of $G_{\text{nms(real)}}$ and $G_{\text{nms(nvs)}}$ because pose-NMS is a non-linear operation.

\subsubsection{Application of clustering and top-grasp filtering}
\label{subsubsection_clustering}
\begin{figure}[htbp]
    \centering
    \resizebox{0.7\width}{!}{ 
    \begin{tikzpicture}[
        node distance = 0.5cm, auto,
        every node/.style = {rectangle, draw, align=center}
        ]

        \node (input1) {$G_{\text{real}}$};
        \node (input2) at ($(input1.center) + (2cm,0)$) {$G_{\text{nvs}}$};
        \node (A1) [below=of input2] {Retain top 50\%};
        \node (B1) [below=of A1] {Cluster: translation distance $0.05m$, rotation distance $10^{\circ}$};
        \node (C1) [below=of B1] {In every cluster, retain best grasp from viewpoint};
        \node (D1) [below=of C1] {In every cluster, retain best grasp};
        \node (output2) [below=of D1] {$G_{\text{cluster(nvs)}}$};
        \node (output1) at ($(output2.center) + (-3cm,0)$) {$G_{\text{cluster(real)}}$};

        \draw[->] (input1) -- (A1);
        \draw[->] (A1) -- (B1);
        \draw[->] (B1) -- (C1);
        \draw[->] (C1) -- (D1);
        \draw[->] (D1) -- (output1);

        \draw[->] (input2) -- (A1);
        \draw[->] (D1) -- (output2);

        \node (input3) at ($(input2.center) + (2cm,0)$) {$G_{\text{real+nvs}}$};
        \node (output3) at ($(output2.center) + (3cm,0)$) {$G_{\text{cluster(real+nvs)}}$};

        \draw[->] (input3) -- (A1);
        \draw[->] (D1) -- (output3);

    \end{tikzpicture}
    }
    \caption{Clustering and top-grasp filtering}
    \label{fig:three_chains}
\end{figure}
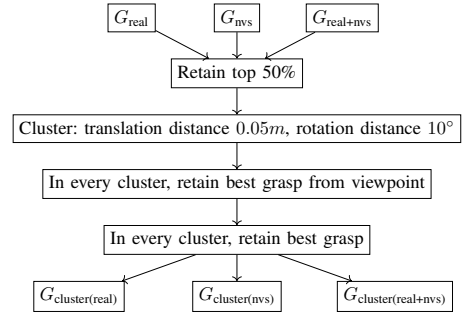

Grasps $G_{\text{real}}$, $G_{\text{nvs}}$, and $G_{\text{real+nvs}}$ are sorted by their predicted scores and the top $50\%$ are retained. These retained grasps are then clustered based on a translation distance threshold of $0.05m$ and rotation distance threshold of $10^{\circ}$. Additionally, every cluster only retains the best grasp from a viewpoint, and subsequently only the top grasp in the cluster is retained. Performing these operations effectively produces only one grasp per cluster, resulting in $G_{\text{cluster(real)}}$, $G_{\text{cluster(nvs)}}$, and $G_{\text{cluster(real+nvs)}}$. Fig.~\ref{fig:three_chains} shows the workflow.

\subsection{Evaluation}
\begin{figure}[htbp]
    \centering
    \includegraphics[width=0.80\columnwidth]{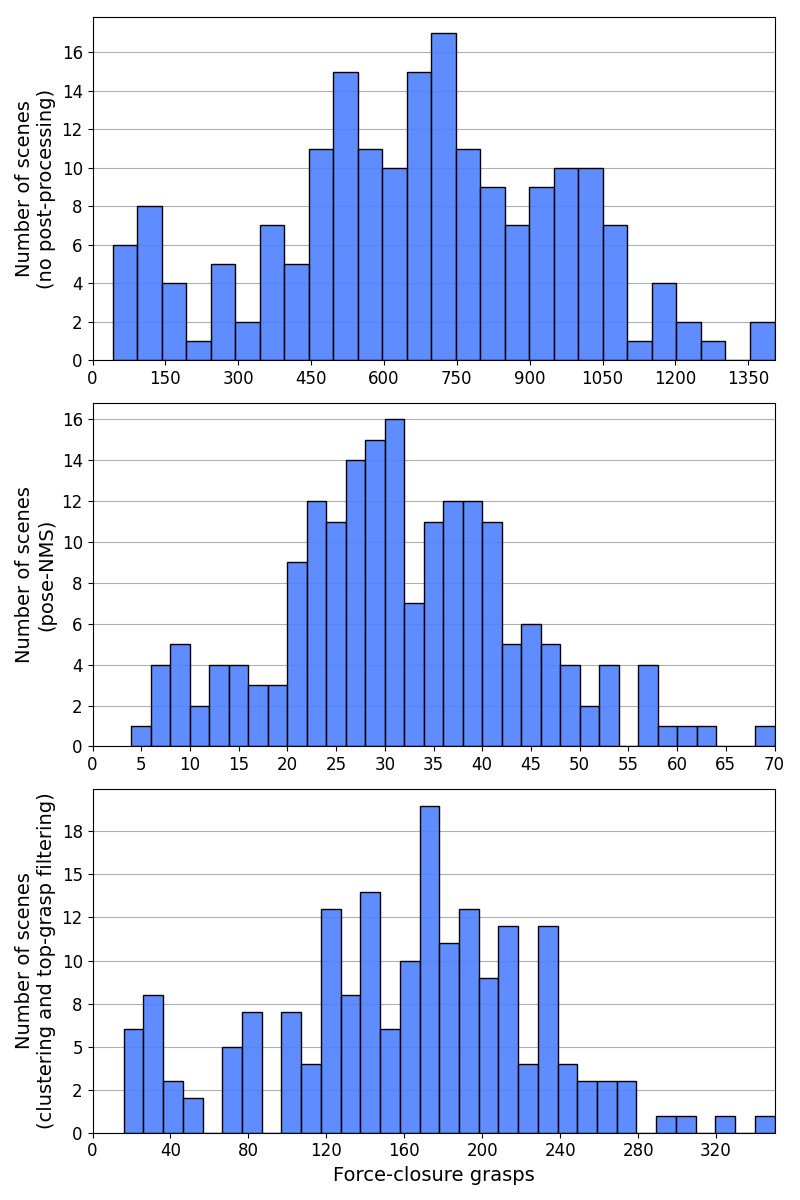}
    \caption{Histogram of force-closure grasps contributed additionally by $N=16$ novel views to the 190 scenes in the Graspnet-1billion dataset}
    \label{fig:fc_grasp_hist}
\end{figure}

The final resulting grasps are checked for force-closure as described in section~\ref{subsection_fc_metric}. Table~\ref{tab:grasp_pose_visualize} shows grasps for an example scene for post-processing branches pose-NMS and clustering and top-grasp filtering. The original grasps $G_{\text{real}}$, $G_{\text{nvs}}$, and $G_{\text{real+nvs}}$ are too many in number resulting in poor visual clarity, and therefore have not been shown. Fig.~\ref{fig:fc_grasp_hist} shows histograms for the number of force-closure grasps contributed by the $N=16$ novel views. The maximum number of grasps occurs without post-processing, as all grasps are retained, unlike the other two post-processing methods: NMS pruning and retention of top grasps based on predicted scores. About 17 out of 190 scenes benefit from approximately 700 force-closure grasps from novel views, with 2 scenes obtaining nearly 1400 grasps each.

Pose-NMS, which merges closely spaced grasps, results in a greater reduction compared to clustering and top-grasp filtering, as evidenced in Table~\ref{tab:grasp_pose_visualize} showing grasps computed on real, novel, and real and novel views for an example scene. For pose-NMS, the largest fraction of scenes gained more than 30 grasps from novel views, while clustering and top-grasp filtering resulted in the largest fraction of scenes gaining close to 180 grasps.

\begin{figure}[htbp]
    \centering
    \includegraphics[width=0.80\columnwidth]{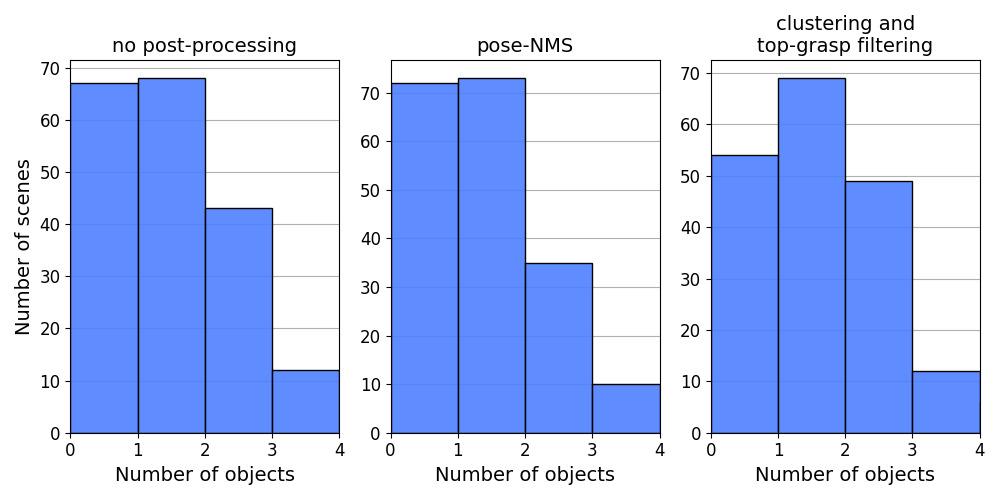}
    \caption{Histogram of number of objects contributed additionally by $N=16$ novel views to the grasp coverage of 190 scenes in the Graspnet-1billion dataset}
    \label{fig:gc_hist}
\end{figure}

Fig.~\ref{fig:gc_hist} presents histograms illustrating how novel perspectives enhance grasp coverage by providing grasp poses for up to four objects that did not have grasp poses from the original views. Most of the scenes got 1 or 2 objects from the novel views, thereby establishing a positive impact of novel views in increasing grasp coverage. This impact is also visible in the grasp poses shown in Table~\ref{tab:grasp_pose_visualize}.

\begin{table}[htbp]
    \caption{Example force-closure grasp poses} 
    \label{tab:grasp_pose_visualize}
    \centering
    \resizebox{0.80\width}{!}{ 
    \begin{tabular}{|>{\centering\arraybackslash}m{0.07\columnwidth}|>{\centering\arraybackslash}m{0.4\columnwidth}|>{\centering\arraybackslash}m{0.4\columnwidth}|}
        \hline
        Views & pose-NMS & Clustering and top-grasp filtering \\
        \hline
        Real & 
        \parbox[c]{0.4\columnwidth}{
            \centering
            \vspace{3pt} 
            Grasp coverage: $77.78 \%$ \\ (7 out of 9 objects) \\
            \includegraphics[width=0.4\columnwidth]{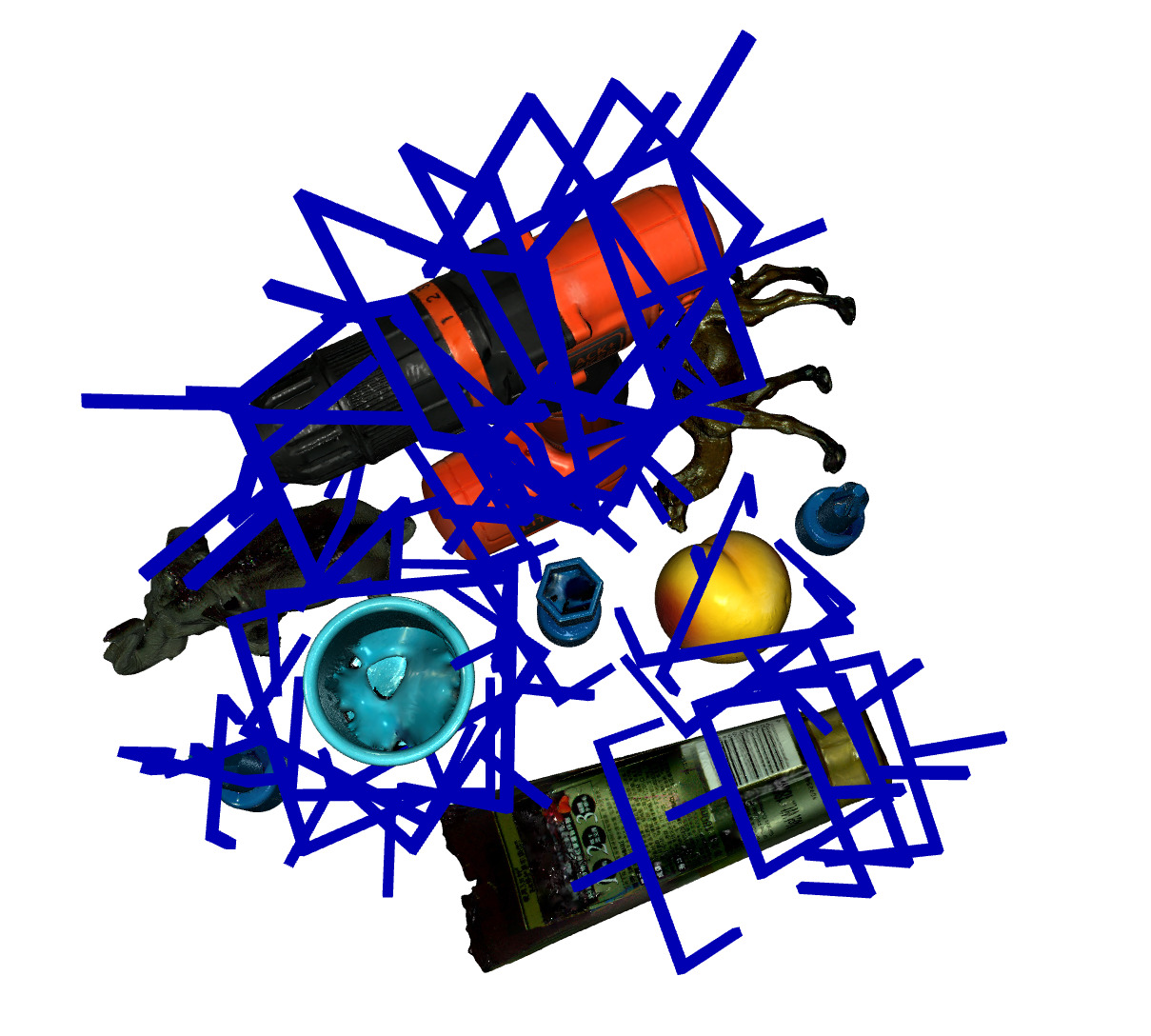}
        } &
        \parbox[c]{0.4\columnwidth}{
            \centering
            \vspace{3pt} 
            Grasp coverage: $44.44 \%$ \\ (4 out of 9 objects) \\
            \includegraphics[width=0.4\columnwidth]{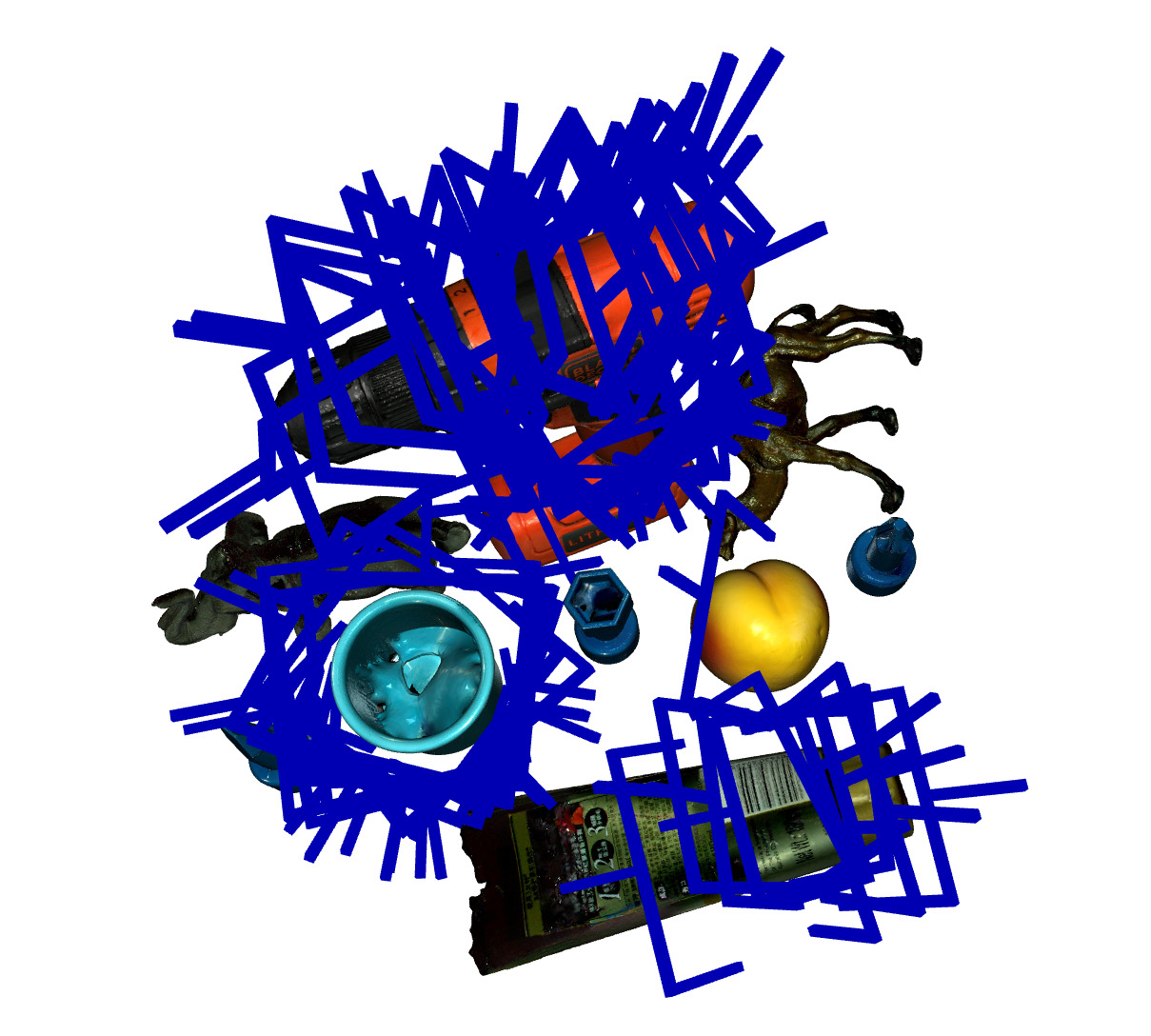}
        } \\
        \hline
        Novel & 
        \parbox[c]{0.4\columnwidth}{
            \centering
            \vspace{3pt} 
            Grasp coverage: $100 \%$ \\ (9 out of 9 objects) \\
            \includegraphics[width=0.4\columnwidth]{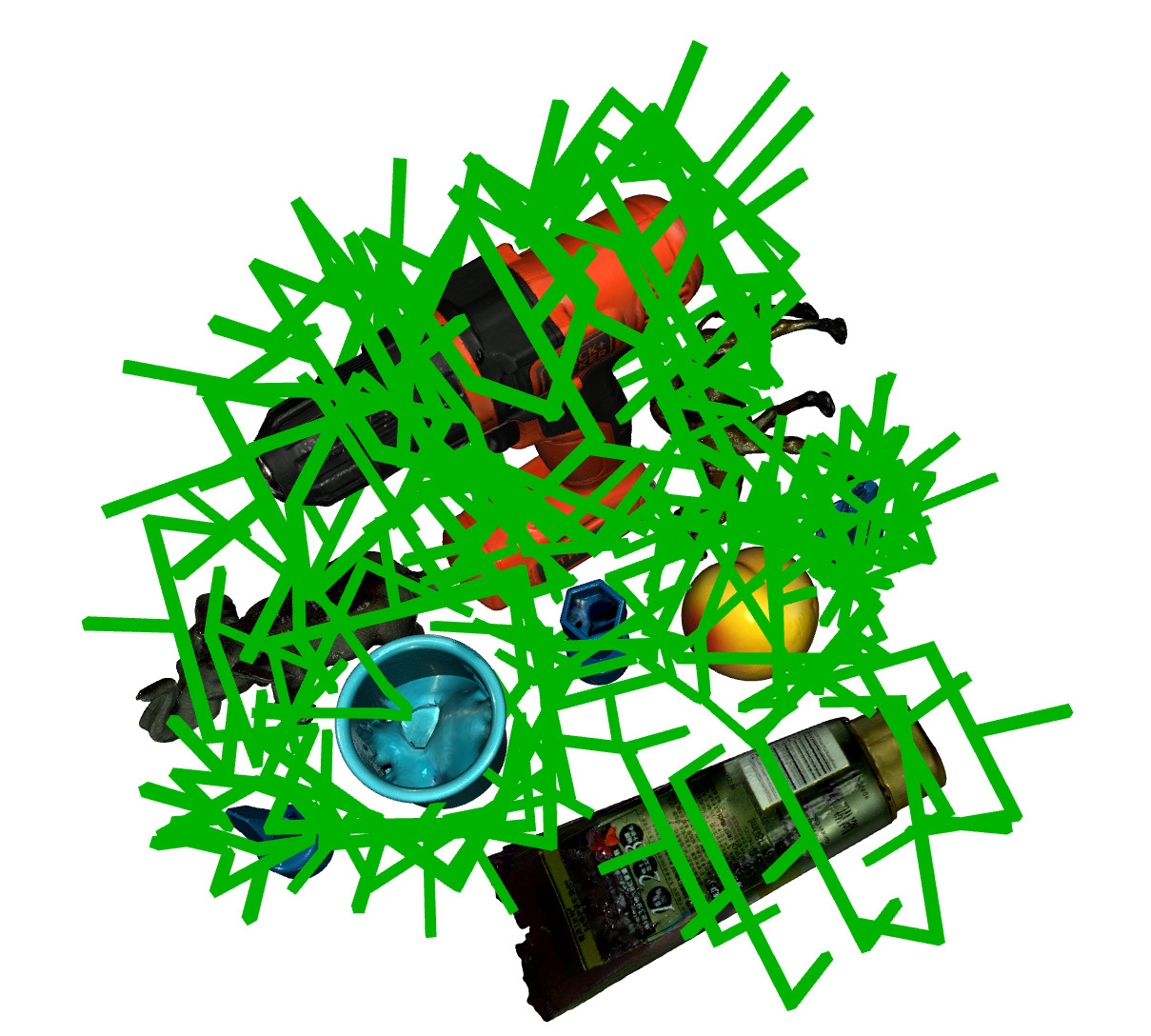}
        } &
        \parbox[c]{0.4\columnwidth}{
            \centering
            \vspace{3pt} 
            Grasp coverage: $66.67 \%$ \\ (6 out of 9 objects) \\
            \includegraphics[width=0.4\columnwidth]{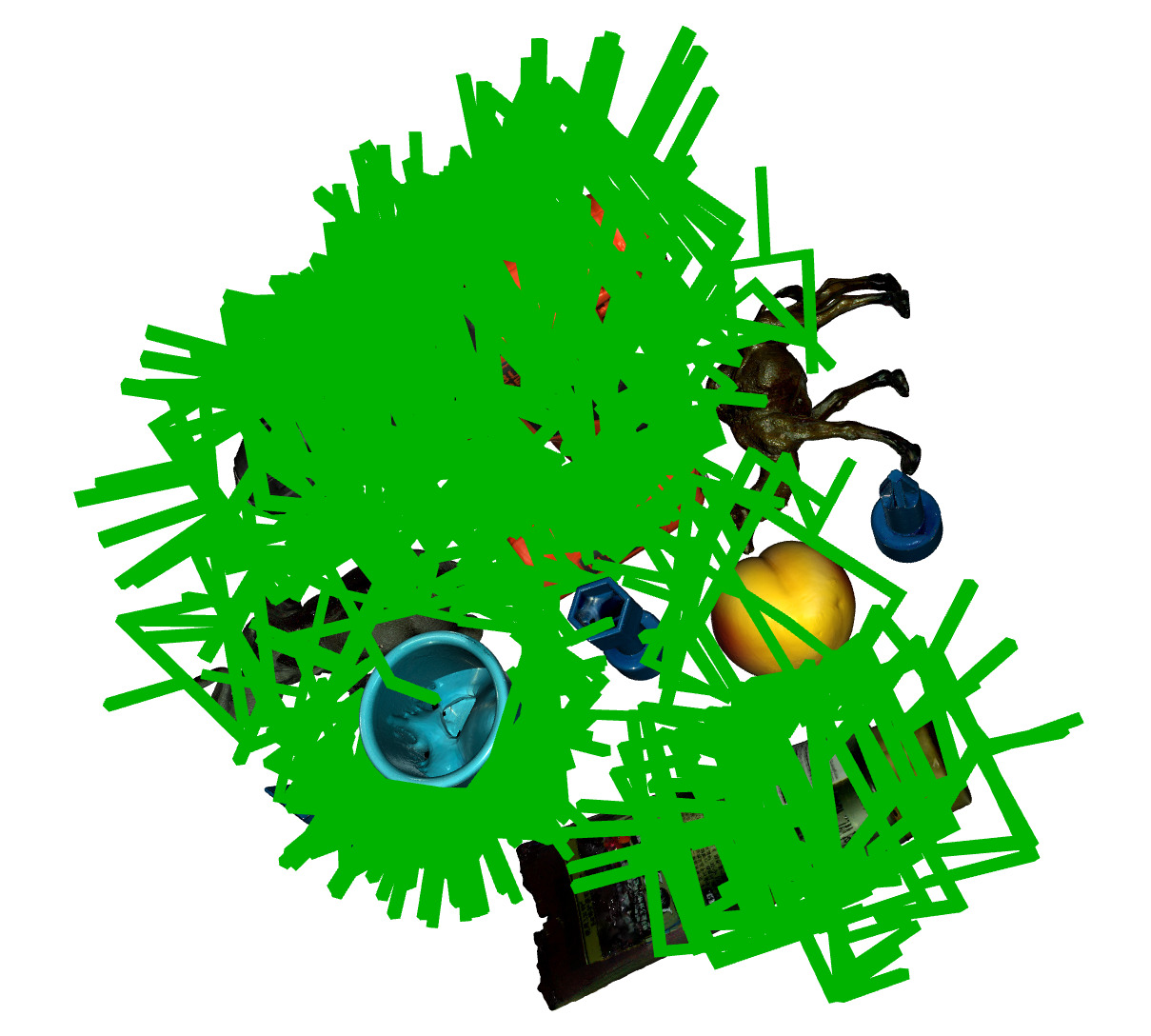}
        } \\
        \hline
        Real + novel & 
        \parbox[c]{0.4\columnwidth}{
            \centering
            \vspace{3pt} 
            Grasp coverage: $100 \%$ \\ (9 out of 9 objects) \\
            \includegraphics[width=0.4\columnwidth]{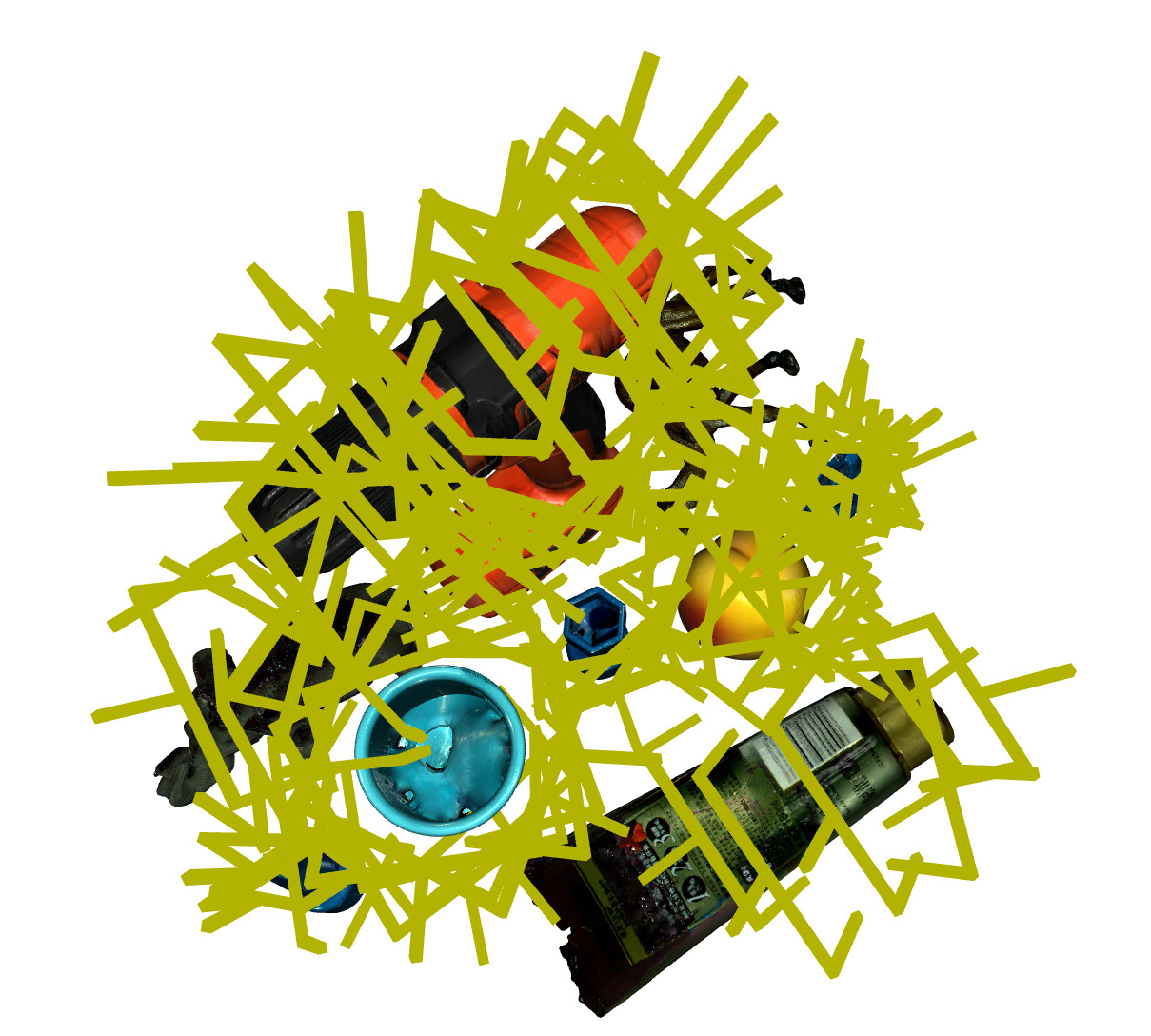}
        } &
        \parbox[c]{0.4\columnwidth}{
            \centering
            \vspace{3pt} 
            Grasp coverage: $66.67 \%$ \\ (6 out of 9 objects) \\
            \includegraphics[width=0.4\columnwidth]{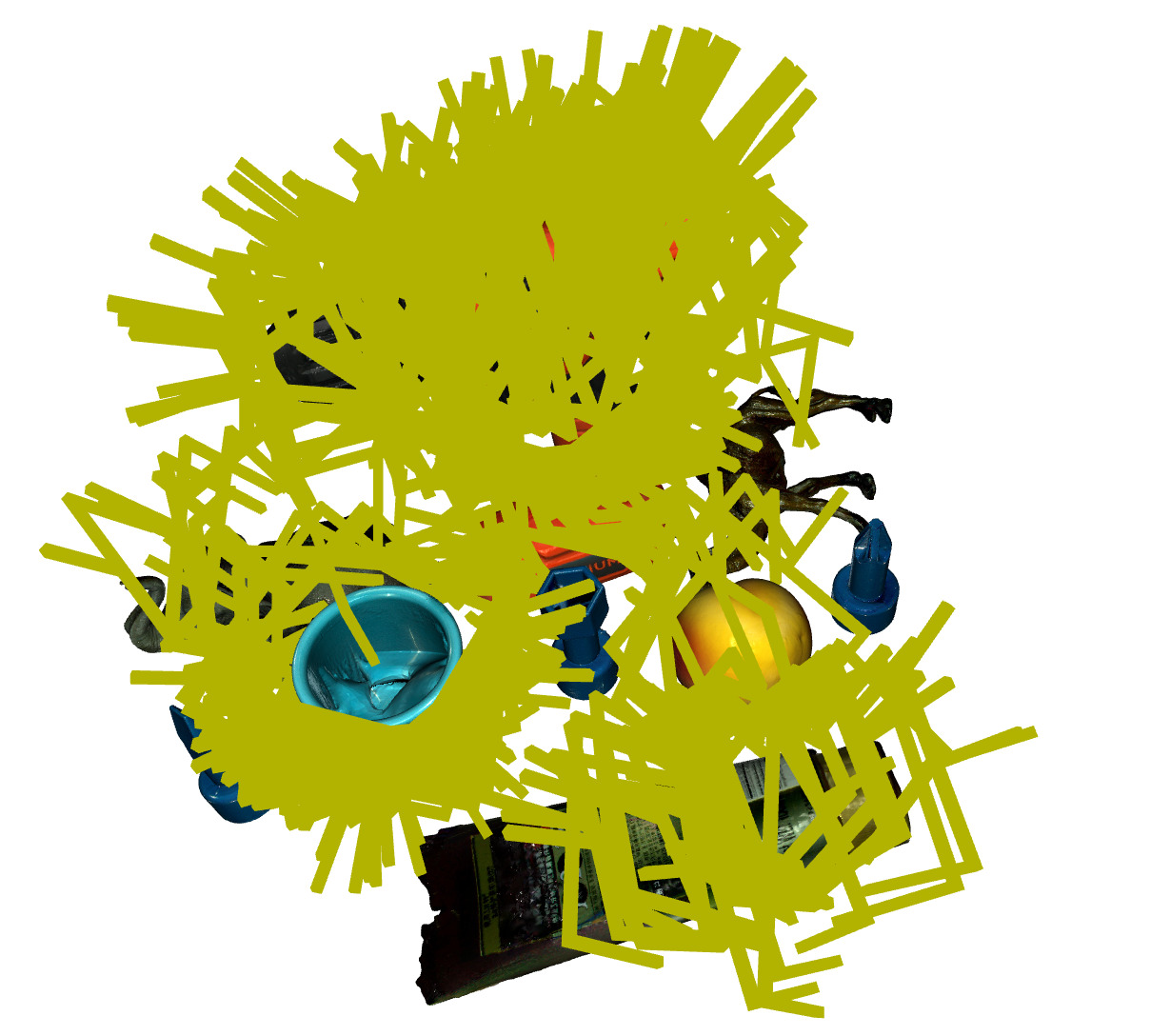}
        } \\
        \hline
    \end{tabular}
    } 
\end{table}



While our evaluation indicates a positive impact of novel-view-synthesis on the number of force-closure grasps, we note two more complex factors:
\begin{itemize}
  \item A high number of force-closure grasps does not come with the guarantee that all of them can be executed on the real robot. Factors that can preclude a successful grasp execution could be one or a combination of unreachability, collision, and a risk of disturbing another object causing the scene representation to become stale.
  \item The grasp coverage percentage is the most optimistic upper-bound. If the only grasp pose or all the grasp poses associated with an object in the scene cannot be executed, that is one less object in the scene that can be grasped.
\end{itemize}

\section{Conclusion}
\label{section_conclusion}
Presented results from experiments run on the Graspnet-1billion dataset indicate that novel views increase the total number of force-closure grasps available for the robot and enable the inference of grasp poses for objects that lacked associated grasps in the real views.

These results require verification on a real robot. Reducing the number of real viewpoints to as few as one~\citep{yu2021pixelnerf, lin2023vision, jang2023nvist}, finding a strategy to get the best novel viewpoints~\citep{lee2022uncertainty, chen2024gennbv}, and improving the grasp extraction process from radiance fields can be promising future directions.

\section*{Acknowledgments}
This work was partially supported by the Wallenberg AI, Autonomous Systems and Software Program (WASP) funded by the Knut and Alice Wallenberg Foundation.

\bibliographystyle{plainnat}
\bibliography{references}

\end{document}